\documentclass[lettersize,journal]{IEEEtran}
\usepackage{algorithmicx}
\usepackage{algpseudocode}  
\usepackage{amsmath,amsfonts}
\usepackage{array}
\usepackage[caption=false,font=normalsize,labelfont=sf,textfont=sf]{subfig}
\usepackage{textcomp}
\usepackage{stfloats}
\usepackage{url}
\usepackage{verbatim}
\usepackage{graphicx}
\usepackage{cite}
\usepackage{CJK}
\usepackage{amsmath}
\usepackage{indentfirst}


\usepackage[linesnumbered,ruled,vlined]{algorithm2e}
\usepackage{color}
\usepackage{bm}

\usepackage{arydshln}
\usepackage{booktabs}
\usepackage{multirow}
\usepackage{threeparttable}

\begin{document}

\title{FastBVP-Net: a lightweight pulse extraction network for measuring heart rhythm via facial videos}
\author{Jialiang Zhuang, Yuheng Chen, Yun Zhang, Xiujuan Zheng 
\thanks{Jialiang Zhuang, Yuheng Chen and Xiujuan Zheng, are all with College of Electrical Engineering, Sichuan University, Chengdu, Sichuan, China. (Corresponding author: Xiujuan Zheng, xiujuanzheng@scu.edu.cn )}
\thanks{Yun Zhang, is with School of Information Science and Technology, Xi'an Jiaotong University, Xi’an, China}}

% \markboth{Journal of \LaTeX\ Class Files,~Vol.~14, No.~8, August~2021}


\maketitle
\begin{abstract}
Remote photoplethysmography (rPPG) is an attractive camera-based health monitoring method that can measure the heart rhythm from facial videos. Many well-established deep-learning models have been reported to measure heart rate (HR) and heart rate variability (HRV). However, most of these models usually require a 30-second facial video and enormous computational resources to obtain accurate and robust results, which significantly limits their applications in real-world scenarios. Hence, we propose a lightweight pulse extraction network, FastBVP-Net, to quickly measure heart rhythm via facial videos. The proposed FastBVP-Net uses a multi-frequency mode signal fusion (MMSF) mechanism to characterize the different modes of the raw signals in a decompose module and reconstruct the blood volume pulse (BVP) signal under a complex noise environment in a compose module. Meanwhile, an oversampling training scheme is used to solve the over-fitting problem caused by the limitations of the datasets. Then, the HR and HRV can be estimated based on the extracted BVP signals. Comprehensive experiments are conducted on the benchmark datasets to validate the proposed FastBVP-Net. For intra-dataset and cross-dataset testing, the proposed approach achieves better performance for HR and HRV estimation from 30-second facial videos with fewer computational burdens than the current well-established methods. Moreover, the proposed approach also achieves competitive results from 15-second facial videos. Therefore, the proposed FastBVP-Net has the potential to be applied in many real-world scenarios with shorter videos.
\end{abstract}

\section{Introduction}
Blood volume pulse (BVP) is an essential physiological signal with great potential in healthcare monitoring. Although many excellent works have proven that the contact-based method can comprehensively and accurately measure cardiac activities, the sensors attached to the human body will bring discomfort and inconvenience. Therefore, in recent years, many remote methods have been proposed for detecting cardiopulmonary signals using video or microwave sensing. In this field, remote Photoplethysmography (rPPG) is a fast-growing technique to measure the heart rhythm from facial videos. In 2008, the rPPG technique was validated to measure the heart rhythm with normal ambient light using the fast Fourier transform combined with a band-pass filter~\cite{40Green2008}. Then, according to the physical principle of rPPG~\cite{44W2017}, the blind source separation (BSS) methods, such as independent component analysis (ICA)~\cite{15BS2006,5Poh2010}, and principal component analysis (PCA)~\cite{20Eduardo2012,26Tang2018}, were used to derive effective signals to measure heart rhythm in the beginning. Then, the Eulerian video magnification (EVM)~\cite{Wu12Eulerian} and the chromaticity-based color space projection (CHROM)~\cite{36Chrom2013} were proposed separately to enhance the hidden information in the videos. As a result, they are usually adopted as the practical pre-processing step to measure heart rhythm~\cite{Qiu2019TMM,Zheng2022,Yang2022THMS}. Moreover, time-domain adaptive filtering~\cite{4Li20142014} and denoising methods~\cite{21Zhang2015,22Murthy2015} are also adopted to eliminate the disturbance for estimating heart rate from facial videos. Nevertheless, these unsupervised signal processing algorithms are limited by poor generalization, making them difficult to deal with the noise caused by the complicated changing ambient lights and the movements.\par
In this case, deep learning is widely considered a better choice to overcome these problems. The powerful semantic feature extraction capability of neural networks, which can effectively learn and deal with complex noise conditions, has led to many deep-learning approaches and achieved great success for heart rhythm estimations in the past five years~\cite{8Niu2018,13DeepPhy2018,14Yu2019}. After obtaining the BVP signals from facial video, the instantaneous heart rate (HR) can be easily calculated according to one cardiac cycle signal. However, the estimated HR could have a significant deviation, while the BVP signals derived by rPPG are contaminated by varied noise and artifacts. Moreover, measuring average HR and heart rate variability (HRV) requires several cycles of stable BVP signals. Therefore, the network must be highly robust to the complex noise for extracting BVP signals, especially when using shorter videos. Furthermore, the extracted BVP signals should be clean and stable enough for accurately measuring HR and HRV. For this purpose, most current networks calculate the average heart rate of couples of video clips truncated from a 30-second facial video~\cite{7DualGANJB,27CVD2020}. However, this repeat computation process is time-consuming and inflexible. Meanwhile, deep-learning algorithms are easily over-fitting because of the imbalance and diversity of individual heart rate distribution under varied lighting conditions.\par
In this case, this study proposes a lightweight pulse extraction network named FastBVP-Net to derive the BVP signals from facial videos. The FastBVP-Net uses a multi-frequency mode signal fusion (MMSF) mechanism, which inherits the idea of the empirical mode decomposition, to improve the robustness of the interference noise. MMSF mechanism consists of a decompose module and a compose module. The decompose module helps the network model the characteristics of the raw signals in multi-frequency bands. The compose module adopts the temporal multiscale convolution and the spectrum-based attention mechanism to capture heart rhythm with limited computational resources. In addition, an oversampling training scheme is proposed to solve the potential over-fitting problem caused by the limitations of the datasets. The details of the proposed FastBVP-Net are described in the following sections.
\section{Related work}
\subsection{Unsupervised signal-processing approaches for remote photoplethysmography}
In the past decade, many studies have explored unsupervised signal-processing approaches to extract BVP signals or measure the heart rhythm from facial videos~\cite{44W2017}. It is proved that the blind source separation (BSS) methods can effectively extract BVP signals while they are still sensitive to movement and illumination variation. In this condition, the combined empirical mode decomposition (EMD) and BSS methods are explored to extract the heart rhythm components of noisy BVP signal efficiently~\cite{19sun2012, MotinJBHI2018}. Furthermore, Cheng et al.~\cite{ChengJBHI2017} propose an illumination-robust framework using joint blind source separation (JBSS) and ensemble empirical mode decomposition (EEMD) to evaluate HR from the facial videos effectively. However, these EMD-based algorithms exhibit a signal aliasing problem that may affect the accuracy of BVP signal extraction. In this case, the variable mode decomposition (VMD) algorithm was proposed to adaptively decompose the intrinsic modes of the noisy pulse signals~\cite{DAS2022}. Moreover, a recent study indicated the feasibility of using adaptive chirp model decomposition to directly model the BVP signals in the time-frequency domain with less prior information~\cite{Zheng2022}. From these previous studies, we can find that the mode decomposition helps eliminate the influences from the varied illuminate and
movements.
\subsection{Deep-learning models for remote photoplethysmography}
Extracting physiological information via deep learning has come a long way, divided into two main categories, heart rate calculation and BVP signal reconstruction. Radim et al.~\cite{10Ramjn2014} used 3D convolution networks to process the video signal directly to estimate HR values. In previous work~\cite{24RhythmNet2020}, a spatial-temporal representation was designed as the input of CNN, which simultaneously contains temporal and spatial features. As a result, the stability and accuracy of heart rate estimation were significantly improved. Although these algorithms have excellent results in HR estimation, they missed a lot of cardiopulmonary information without extracting BVP signals. In this condition, a 3D spatial-temporal network~\cite{25yuPhysNet2019} is proposed to derive the BVP signal from facial videos. It first offered Pearson correlation coefficients as the signal reconstruction's loss function, significantly improved network performance. Then, a video enhancement network is proposed to recover rPPG signals from highly compressed facial videos~\cite{14Yu2019}. To deal with the less-constrained scenarios, Yu et al.~\cite{1yuAutoHR2020} use a powerful searched backbone with novel temporal difference convolution to capture intrinsic features of BVP signals between video frames. Niu et al.~\cite{27CVD2020} designed a cross-verified feature separation strategy to separate physiological and non-physiological features and then performed robustly multitask heart rhythm measurements using the extracted physiological features. Lu et al.~\cite{7DualGANJB} proposed a remote physiological measurement algorithm called DualGAN, jointly modeling the BVP predictor and noise distribution to suppress the noise mixed with the physiological information. The deep learning methods with advanced network architecture and long-enough face videos can be robust to complex noise.
\subsection{Attention mechanism}
Recently, there have been several works to incorporate attention processing to improve the performance of convolutional neural networks (CNNs), which helps the network perform well with noisy inputs. Many existing well-established attention mechanisms modeling in multiple dimensions have achieved tremendous success in a classification tasks. Among them, the non-local network~\cite{32Wang_2018} pioneered the algorithm of aggregating global context to each query location to capture long-term dependencies. Cao et al.~\cite{31Cao_2019} proposed to apply a simplified design similar to the squeezed excitation network (SENet)~\cite{33hu2018} to the above network structure, which can design a three-step generic framework for modeling global context more effectively. In addition, Woo et al.~\cite{30Woo_2018} created a lightweight generic module that performs adaptive feature refinement along the channel and spatial dimensions. The core constructs in the well-known Transformer~\cite{29vaswani2017} network is the multi-headed self-attentive mechanism, which shows that the self-attentive tool for extracting global features has excellent potential to improve algorithmic capabilities. 
\section{Method}
\begin{figure*}[ht]
	\centering
        \includegraphics[width=0.9\linewidth]{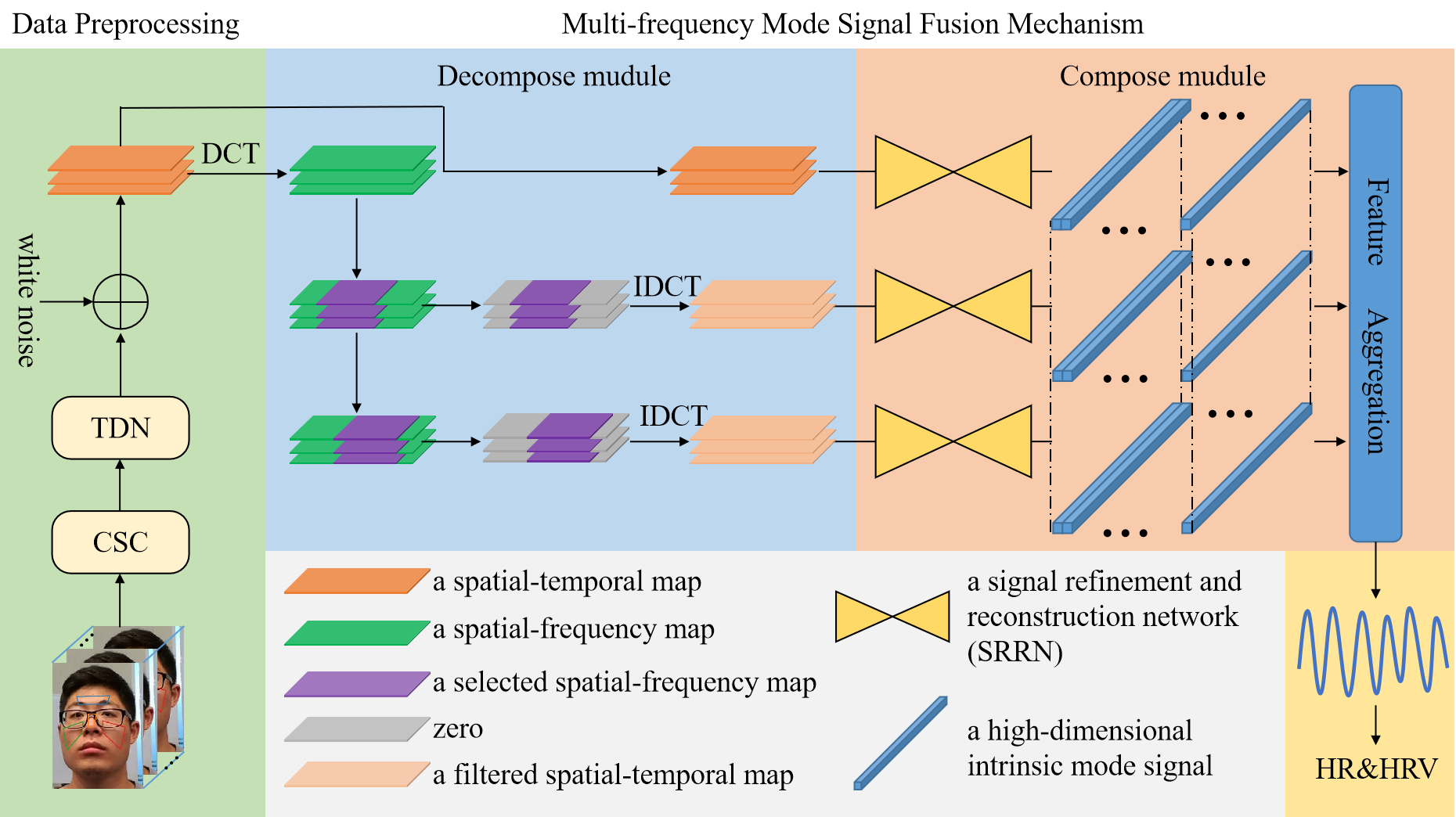}
	  \caption{Workflow of the proposed FastBVP-Net. The input video clips are cut into four regions based on the facial landmarks and pre-processed by color space conversion (CSC) and time-domain normalization (TDN). Then, the white noise is added to generate the spatial-temporal map. The multi-frequency mode signal fusion mechanism consists of a decompose module and a compose module. First, the decompose module extracts multi-frequency mode components from the spatial-temporal map in frequency domains, obtaining multi-band signals. Then, in the compose module, the spatial-temporal map and the multi-band signals are input into the signal refinement and reconstruction network (SRRN) to get the intrinsic mode signals and reconstruct the blood volume pulse (BVP) signals. Finally, the heart rate (HR) and heart rate variability (HRV) can be calculated based on the reconstructed BVP signals.}
	  \label{workflow}
\end{figure*}

In this section, we explore a lightweight pulse extraction network, FastBVP-Net, for quickly measuring heart rhythm via facial videos. After data preprocessing, the FastBVP-Net uses the multi-frequency mode signal fusion mechanism to derive the BVP signals and then calculate the physiological parameters. The workflow of the proposed method is shown in Fig.~\ref{workflow}.
\subsection{Data preprocessing}
The physiological information reflecting the cardiac cycles is weak in facial videos and is usually contaminated by the noise caused by lighting variation and head movements. Therefore, it is crucial to enhance the physiological information in facial videos.\par
The step of data preprocessing first compresses the input video clips into an initial spatial-temporal map. Specifically, the facial landmarks are detected on the $i^{th}$ frame of the facial videos. The average pixel values of the four regions are obtained by cutting the facial regions with these landmarks. Then the color space conversion (CSC) is used to convert the initial spatial-temporal map to the modified YUV color space. The modified YUV color space gives more attention to brightness dimension features in color space and reduces the noise caused by the difference in ambient light at different locations on the face. The modified YUV color space can be formulated as~\ref{eq1}.\par
\begin{footnotesize}  
\begin{equation}
\left[\begin{array}{l}
Y_m \\
U_m \\
V_m
\end{array}\right]=\left[\begin{array}{ccc}
0.299 & 0.587 & 0.114 \\
-0.169 & -0.331 & 0.5 \\
0.5 & -0.419 & -0.081
\end{array}\right]\left[\begin{array}{c}
R\\
G\\
B
\end{array}\right]
\label{eq1}
\end{equation}
\end{footnotesize} 
\noindent
where $Y_m$, $U_m$ and $V_m$ are for the Y, U, and V channels in the modified YUV color space, respectively.\par
Then, the time-domain normalization module (TDN) is performed to obtain the normalized spatial-temporal map. TDN is proven to be effective in helping networks recover physiological signals accurately in the face of large amounts of noise interference~\cite{27CVD2020}. Finally, we add the white noise to the normalized spatial-temporal. The white noise effectively simulates various potential noisy signals caused by extra disturbances, such as harsh ambient light conditions and motion expressions in the actual scenes.\par
\subsection{Multi-frequency mode signal fusion mechanism}
Blind source signal separation algorithms have been shown to efficiently filter noise and extract BVP signals from reflected light signals from faces videos~\cite{20Eduardo2012}. Various time-domain filters and frequency-domain filters can be considered signal separators. Most blind source signal separation algorithms use unsupervised methods, such as independent component analysis, which decomposes the received mixed signal into several separate components as an approximate estimate of the source signal. Moreover, blind source separation algorithms, including PCA and ICA, are only suitable for processing linearly correlated mixed signals. However, remote photoelectric volume pulse signals are interspersed with complex noise, and blind source separation algorithms can struggle in this situation. Therefore, in the task of BVP extraction, we are faced with an unknown distribution of the composite signal mixed with a large amount of noise, which significantly limits the effectiveness of the blind source signal separation method.\par
In order to handle complex signals with non-linear correlation, we adaptively extract signal features in each frequency band. This paper proposes the multi-frequency mode signal fusion mechanism to break this challenge. Specifically, as shown in Fig.~\ref{workflow}, we first use discrete cosine transform (DCT) and inverse discrete cosine transform (IDCT) to obtain the initial multi-frequency signals in the multi-frequency mode signal decompose module. Then we exploit a signal refinement and reconstruction network (SRRN) to extract multi-scale features from each component of the multi-frequency mode signal, obtaining the high-dimensional intrinsic mode signal. In the multi-frequency mode signal compose module, we extract the BVP signal by fusing the practical components of the high-dimensional intrinsic mode signal.
\subsubsection{Multi-frequency mode signal decompose module}
The reflected light intensity of each face region can illuminate the intravascular substance concentration changes in this region, i.e., the characteristics of the different phases of the cardiac cycle, so that we can consider the one-dimensional signals of each color channel in each facial region over a long period as individual pulse signals. %We perform frequency domain operations on these initial pulse wave signals.\par
In order to extract multi-frequency mode components from the spatial-temporal map in the frequency domain, we set $K$ frequency bands of interest and $I$ facial regions, and the detailed process is shown in Algorithm~\ref{algorithm1}.\par
First, we extract the time-domain signal $f_{i}(n)$ on each facial region from the spatial-temporal map. The discrete cosine transform is then applied to each input time-domain signal $f_{i}(n)$ to get $F_{i}(u)$.\par
Based on the boundary defined by $K$ frequency bands of interest, we cut part of the frequency spectrum and do zero-padding to get an adjusted frequency spectrum $F^{\prime}_{i,k}(u)$, which is then transformed to the filtered time domain signal $f^{\prime}_{i,k}(n)$ by inverse discrete cosine transform.\par
Finally, we concatenate filtered time domain signals $f^{\prime}_{i,k}(n)$ of each facial region and obtain multi-band signals $f^{\prime}_{k}(n)$ of each frequency band.
\begin{algorithm}  
  \caption{Multi-frequency mode signal decompose module} \label{algorithm1} 
  \KwIn{Spatial-temporal map \textbf{M}, set of frequency bands $B_{k}$, set of facial regions $R_{i}$, where $k = 1, ..., K$, and $i = 1, ..., I$}  
  \KwOut{Multi-band signals $f^{\prime}_{k}(n)$} 
  \For{$i^{th}$ facial region in \rm{\textbf{M}}}    
  { Get the time domain signals $f_{i}(n)$\;
  Do discrete cosine transform to obtain frequency spectrum $F_{i}(u)$\;
    \For{$k^{th}$ frequency band $B_{k}$}  
    {  
      Cut part of the frequency band and do zero-padding to get adjusted frequency spectrum $F^{\prime}_{i,k}(u)$\;
	  Use inverse discrete cosine transform to obtain filtered signal $f^{\prime}_{i,k}(n)$\;}  
    Concatenate $f^{\prime}_{i,k}(n)$ of each facial region to obtain multi-band signal $f^{\prime}_{k}(n)$ of each frequency band \;}  
  return $f^{\prime}_{k}(n)$\;  
\end{algorithm}
\subsubsection{Multi-frequency mode signal compose module}
In order to extract the clean BVP signal from multi-frequency mode signals, a signal refining and reconstruction network is proposed in this paper, including a signal refinement sub-network and signal reconstruction sub-network.\par
\begin{figure*}[htp]
	\centering	\includegraphics[width=0.8\linewidth]{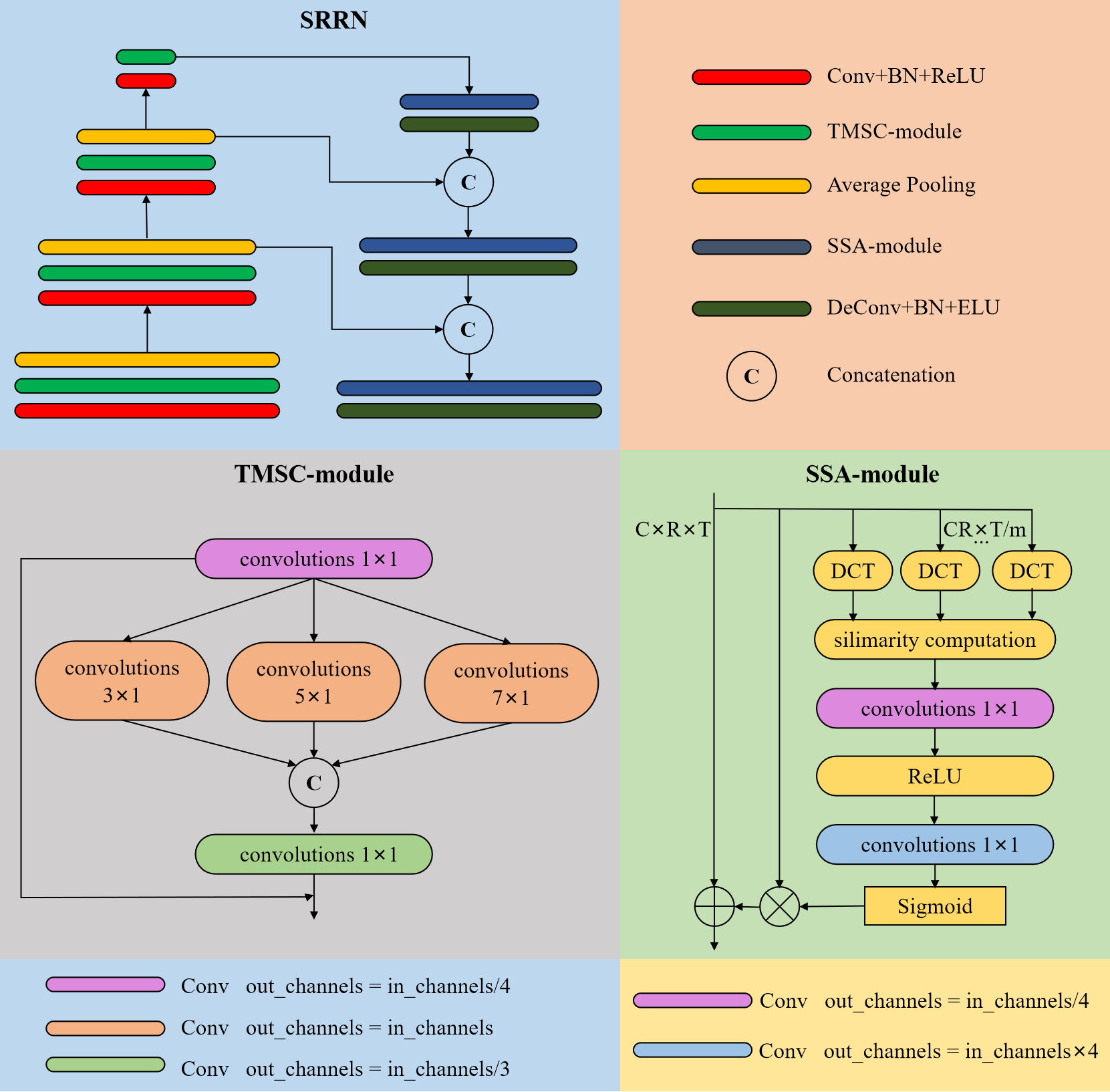}
	  \caption{Architecture of the signal refinement and reconstruction network (SRRN). The signal refinement sub-network consists of four tandem modules, each consisting of a solution layer, a Batch Normalization (BN) layer, a Relu layer, a temporal multiscale convolution module (TMSC-module), and a pooling layer. TMSC-module uses three kinds of convolution kernels ($3\times1$, $5\times1$, $7\times1$) to extract multiscale time-domain features. The signal reconstruction sub-network consists of three tandem modules, each consisting of a deconvolution (DeConv) layer, a BN layer, an ELU layer, and a spectrum self-attention module (SSA-module). The SSA-module inherits the scaled dot-product attention from the transformer, which calculates the similarity of spectral features at different periods in each dimension. Finally, two convolution layers are used to obtain the global perceptual field.}
	  \label{SRRN}
\end{figure*}

We design a signal refinement sub-network to extract signal features at multiple resolutions. The spatial-temporal map and the multi-band signals are fed into the network, and a corresponding high-dimensional frequency signal is the output. The pipeline of the signal refinement sub-network is designed as shown in Figure~\ref{SRRN}, which consists of a temporal multi-scale convolution module (TMSC-module), a convolution layer, an activation layer, a BN layer, and a pooling layer. It aims to gradually map the original signal to relatively pure potential stream shapes at lower resolutions, enhancing the network's ability to filter out noisy signals adaptively.\par
In the signal refinement sub-network, the TMSC-module is the main structure that can improve the network's ability while keeping the computational budget constant. Besides, to capture multi-scale features in the time domain, we set the convolution kernel sizes of the TMSC-module to $3\times1, 5\times1, 7\times1$ similar to the settings in the previous work~\cite{43Inception}. Moreover, we concatenate the output of the three convolution blocks for fusing multi-scale features and obtain a better representation to highlight the physiological information.\par
As shown in Figure~\ref{SRRN}, the signal reconstruction sub-network, including the spectrum self-attention module (SSA-module), a deconvolution layer, an activation layer, and a BN layer, which uses the multi-scale features in the feature extraction stage to perform signal reconstruction. It aims to improve the signal-to-noise ratio and get the precise position of each peak of the reconstructed BVP signal.\par
We assume that the frequency domain features of different periods should be similar in a signal waveform. If there is a significant difference in the waveform features in a certain period compared with others, it confirms the presence of significant noise signals at this time. We can get the self-attention weights of each filter through this mechanism and then put these weights into the channel attention module to globally group the individual filter features. This process can further reduce the noise.\par
In this study, the proposed spectrum-based attention mechanism can be abstracted into three steps: (a) The signal is segmented in the time domain and then subjected to a discrete cosine transform to obtain the frequency features. (b) The scaled dot-product attention in transformer~\cite{29vaswani2017} is adopted in the similarity computation. (c) After obtaining the similarity of each dimension, we put them into a global representator composed of convolution layers, which aims to perceive the characteristics of other dimensions and obtain the global receptive field. Our approach falls under the attention mechanism-based approach. However, with several significant differences compared to existing methods: (1) Different from~\cite{31Cao_2019,32Wang_2018} requiring high-dimensional semantic features as the input, we convert the signal features into a spectrum in the time domain dimension, which is then fed into attention module. (2) Unlike~\cite{33hu2018} let the network automatically learn the weight of each dimension feature, we mainly locate the noise and calculate its weight by looking for the difference of signals in different periods. We fuse multiple high-dimensional intrinsic mode signals to extract the practical component and reconstruct the target BVP signals.\par
\subsection{Oversampling training scheme}
Due to the highly unbalanced distribution of training samples in terms of heart rate, putting all samples directly into the network training will inevitably overfit a specific heart rate range. Therefore, to let the network effectively learn the pulse signals corresponding to all heart rate intervals, we first train the network with all training samples. Then, we use a lower learning rate to update the method with an oversampling training scheme that divides the training samples into several groups according to the heart rate range. In each batch, the number of samples in each group is guaranteed to be in the same proportion, together with the data enhancement described in the step of data pre-processing, to maximize the limited training data set to strengthen the network fitting ability.\par
\section{Experiments and results}
\subsection{Datasets and experimental settings}
\subsubsection{Datasets}
We evaluate our method on three widely-used public datasets for physiological measurement, i.e., UBFC-rPPG, VIPL-HR and MMSE-HR.\par
\textbf{UBFC-rPPG} is a challenging dataset of remote physiological measurements with sunlight and indoor lighting conditions. It contains 42 RGB videos, all captured at 30 frames per second (fps) using a Logitech C920 HD Pro webcam, and corresponding actual BVP signals acquired using a CMS50E. In our experiments, UBFC-rPPG was used in intra-dataset testing and Ablation study. We train the network on the first 30 subjects and test the remaining 12 subjects.\par
\textbf{VIPL-HR} contains 2378 less-constrained RGB videos from 107 individuals, which include a variety of scenarios such as head movements, lighting changes, etc. The frame rate of all videos varies from 25 fps to 30 fps. In this study, we use this dataset for training in cross-dataset testing.\par
\textbf{MMSE-HR} is a large dataset for remote measurement of heart rate. It has 102 videos taken from 40 subjects. In addition, it contains videos of multiple facial expressions and head movements recorded at 25 fps. In this study, we use MMSE-HR dataset for testing in cross-dataset testing.\par
\subsubsection{Evaluation metrics}
For the task of average HR estimation, we use the metrics including the standard deviation of the error (Std), the mean absolute error (MAE), the root mean squared error (RMSE), and the Pearson's correlation coefficients (r). Following the previous study~\cite{26Tang2018,27CVD2020}, we use low frequency (LF), high frequency (HF), and the ratio of low and high frequency (LF/HF) in terms of Std, RMSE, and r for the task of HRV estimation.\par
\subsection{Intra-dataset testing}
\subsubsection{HR estimation on UBFC-rPPG}
We evaluate the average HR estimation on UBFC-rPPG. The popular methods, including unsupervised signal processing methods (POS~\cite{38POS2017}, CHROM~\cite{36Chrom2013} and Green~\cite{40Green2008}) as well as deep-learning based methods (SynRhythm~\cite{9Antony2015}, PulseGAN~\cite{39PulseGAN2021}, DualGAN~\cite{7DualGANJB}) are used for comparison. We directly take the results of these well-established methods from previous work. For the comparative methods, the 30 seconds video is segmented into multiple clips for HR estimation. The average HR value of all clips is defined as the final result for the 30 seconds video. In this study, we reconstruct the BVP signal based on the 30 seconds video and then detect the peaks of the BVP for calculating the average HR. The comparative results are given in Table~\ref{HR-intra}. In the case of HR calculation, the proposed method achieves promising results with an MAE of 0.75 beat per minute (bpm), an RMSE of 1.12 bpm, and an r of 0.99. It outperforms the recent well-established traditional and deep learning methods, except for Dual-GAN~\cite{7DualGANJB}). Most of these comparative methods need to get the average result of multiple overlapping video clips. The proposed method generates a higher quality waveform of BVP signals and obtains a more accurate HR in on calculation for a 30-second video.
\begin{center}
\begin{table}[htb]
%\vspace{-2.5em}
\centering
\caption{\centering{HR estimation results of FastBVP-Net and well-established methods on the UBFC-rPPG dataset.}}\label{HR-intra}
\setlength{\tabcolsep}{2.5mm}{
\begin{tabular}{ccccc}
\toprule
\textbf{Method} & \textbf{MAE$\downarrow$} & \textbf{RMSE$\downarrow$}  & \textbf{r$\uparrow$}\\ 
\midrule
\textbf{POS~\cite{38POS2017}} & $8.35$ & $10.00$ & $0.24$\\
\textbf{CHROM~\cite{36Chrom2013}} & $8.20$ & $9.92$ & $0.27$ \\
\textbf{Green~\cite{40Green2008}} & $6.01$ & $7.87$ & $0.29$\\
\textbf{SynRhythm~\cite{8Niu2018}} & $5.59$ & $6.82$ & $0.72$ \\
\textbf{PulseGAN~\cite{39PulseGAN2021}} & $1.19$ & $2.10$ & $0.98$ \\
\textbf{Dual-GAN~\cite{7DualGANJB}} & \bm{$0.44$} & \bm{$0.67$} & \bm{$0.99$} \\ 
\textbf{FastBVP(30s)} & $\textit{0.75}$ & $\textit{1.12}$ & $\textit{0.99}$ \\
\specialrule{0em}{1pt}{1pt}
\hdashline
\specialrule{0em}{1pt}{1pt}
\textbf{FastBVP(15s)} & $1.70$ & $2.01$ & $0.98$ \\
\bottomrule
\end{tabular}}
\begin{tablenotes}
\footnotesize
\item[*] Notes: Bold indicates best performance, italic indicates second best performance.
\end{tablenotes}
\end{table}
\end{center}
\subsubsection{HRV estimation on UBFC-rPPG}
Following the previous study~\cite{7DualGANJB}, we use the first 30 subjects for training and the remaining 12 for testing. For the task of HRV estimation, we use several well-established methods including POS~\cite{38POS2017}, CHROM~\cite{36Chrom2013}, Green~\cite{40Green2008}, CVD~\cite{27CVD2020}, DualGAN~\cite{7DualGANJB} for comparison. The results of these comparative methods are inherited from the published work~\cite{7DualGANJB}. The comparative results of HRV estimation are shown in Table~\ref{HRV-intra}. We can see that the proposed FastBVP-Net outperforms these well-established methods for HRV estimation under all evaluation metrics. Furthermore, the results in Table~\ref{HR-intra} and Table~\ref{HRV-intra} show that the proposed method can accurately recover the heart rhythm, including HR and HRV, from a facial video.
\begin{center}
\begin{table*}[htbp]
\centering
\caption{\centering{HRV estimation results of FastBVP-Net and well-established methods on the UBFC-rPPG dataset.}}\label{HRV-intra}
\setlength{\tabcolsep}{2mm}{
\begin{threeparttable}
\begin{tabular}{ccccccccccccc}
\toprule
\textbf{Method} && \textbf{LF-(u.n)} &&& \textbf{HF-(u.n)} &&& \textbf{LF/HF} &\\ 
\midrule
       &\textbf{Std$\downarrow$}& \textbf{RMSE$\downarrow$} &\textbf{r$\uparrow$}&\textbf{Std$\downarrow$}&\textbf{RMSE$\downarrow$}  &\textbf{r$\uparrow$}&\textbf{Std$\downarrow$}& \textbf{RMSE$\downarrow$} &\textbf{r$\uparrow$}\\ 
\midrule
\textbf{POS~\cite{38POS2017}}& $0.171$ & $0.169$ & $0.479$ & $0.171$ & $0.169$ & $0.479$  & $0.405$ & $0.399$ & $0.518$  \\
\textbf{CHROM~\cite{36Chrom2013}}& $0.243$ & $0.240$ & $0.159$ & $0.243$ & $0.240$ & $0.159$  & $0.655$ & $0.645$ & $0.226$ \\
\textbf{Green~\cite{40Green2008}}& $0.186$ & $0.186$ & $0.280$ & $0.186$ & $0.186$ & $0.280$  & $0.405$ & $0.399$ & $0.518$  \\
\textbf{CVD~\cite{27CVD2020}}& $0.053$ & $0.065$ & $0.740$ & $0.053$ & $0.065$ & $0.740$  & $0.169$ & $0.168$ & $0.812$  \\
\textbf{Dual-GAN~\cite{7DualGANJB}}& $\textit{0.034}$ & $\textit{0.035}$ & $\textit{0.891}$ & $\textit{0.034}$ & $\textit{0.035}$ & $\textit{0.891}$  & $\textit{0.131}$ & $\textit{0.136}$ & $\textit{0.881}$  \\
$\textbf{FastBVP(30s)}$& \bm{$0.030$} & \bm{$0.030$} & \bm{$0.921$} & \bm{$0.030$} & \bm{$0.031$} & \bm{$0.921$}  & \bm{$0.101$} & \bm{$0.101$} & \bm{$0.895$}  \\
\specialrule{0em}{1pt}{1pt}
\hdashline
\specialrule{0em}{1pt}{1pt}
$\textbf{FastBVP(15s)}$& $0.041$ & $0.042$ & $0.850$ & $0.041$ & $0.042$ & $0.85$  & $0.143$ & $0.143$ & $0.853$  \\
\bottomrule
\end{tabular}
\end{threeparttable}}
\end{table*}
\end{center}
\subsection{Cross-dataset testing}
Besides the intra-dataset tests on the UBFC-rPPG dataset. Following the previous study~\cite{27CVD2020}, we train the proposed method on VIPL-HR dataset and test it on MMSE-HR dataset. The results of the proposed FastBVP-Net and recent well-established methods are given in Table~\ref{HR-cross}, in which the results of Li2014~\cite{41Li2014}, CHROM~\cite{36Chrom2013}, SAMC~\cite{37SAMC2016}, RhythmNet~\cite{24RhythmNet2020}, CVD~\cite{27CVD2020} are from the recent published paper~\cite{7DualGANJB}. For HR estimation in cross-dataset testing, the FastBVP-Net gives out the results with an MAE of 5.6 bpm, an RMSE of 5.75 bpm, and an r of 0.90. From Table~\ref{HR-cross}, it can be found that the proposed FastBVP-Net achieves the best results than the compared well-established methods when using 30s facial videos. Considering the results in Table~\ref{HR-intra} and Table~\ref{HR-cross} together, the proposed FastBVP-Net obtained the second-best performance of HR estimation for intra-dataset testing and the best for cross-dataset testing. Furthermore, since the MMSE-HR is a large dataset with varied movements and uncertainties in the facial videos, the results indicate that the proposed FastBVP-Net generalizes well in unconstrained scenarios.
\begin{center}
\begin{table}[htbp]
\centering
\caption{\centering{HR estimation results of FastBVP-Net and several well-established methods in the cross-dataset testing.}}\label{HR-cross}
\setlength{\tabcolsep}{2.5mm}{
\begin{tabular}{ccccc}
\toprule
\textbf{Method} & \textbf{MAE$\downarrow$} & \textbf{RMSE$\downarrow$}  & \textbf{r$\uparrow$}\\ 
\midrule
\textbf{Li2014~\cite{41Li2014}} & $20.02$ & $19.95$ & $0.38$\\
\textbf{CHROM~\cite{36Chrom2013}} & $14.08$ & $13.97$ & $0.55$ \\
\textbf{SAMC~\cite{37SAMC2016}} & $12.24$ & $11.37$  & $0.71$\\
\textbf{RhythmNet~\cite{24RhythmNet2020}} & $6.98$ & $7.33$  & $0.78$ \\
\textbf{CVD~\cite{27CVD2020}} & $\textit{6.06}$ & $\textit{6.04}$  & $\textit{0.84}$ \\
$\textbf{FastBVP(30s)}$ & \bm{$5.6$} & \bm{$5.75$} & \bm{$0.90$} \\
\specialrule{0em}{1pt}{1pt}
\hdashline
\specialrule{0em}{1pt}{1pt}
\textbf{FastBVP(15s)} & $6.57$ & $6.61$ & $0.86$ \\
\bottomrule
\end{tabular}}
\end{table}
\end{center}
\subsection{Ablation study}
We also conduct the ablation study to evaluate the effectiveness of the modified YUV color space, MMSF mechanism, and oversampling training strategy adopted in the proposed method for HR estimation on the UBFC-rPPG dataset. All the comparative results are illuminated in Table~\ref{ablation}.
\begin{center}
\begin{table}[htb]
\centering
\caption{\centering{Ablation study of the FastBVP-Net for HR estimation on UBFC-rPPG}}\label{ablation}
\setlength{\tabcolsep}{2.5mm}{
\begin{tabular}{ccccc}
\toprule
\textbf{Method} & \textbf{MAE$\downarrow$} & \textbf{RMSE$\downarrow$} & \textbf{r$\uparrow$}\\ 
\midrule
\textbf{Without modified YUV} & $2.35$ & $4.56$ & $0.91$ \\
\textbf{Without MMSF} & $2.05$ & $4.01$ & $0.96$ \\
\textbf{Without TMSC} & $0.85$ & $1.24$ & $0.98$ \\
\textbf{Without SSA} & $0.92$ & $1.35$ & $0.98$\\
\textbf{Without OSS} & $1.55$ & $3.13$ & $0.92$\\
\textbf{FastBVP} & \bm{$0.75$} & \bm{$1.12$}& \bm{$0.99$}  \\
\bottomrule
\end{tabular}}
\end{table}
\end{center}
\subsubsection{Effectiveness of the modified YUV color space}
As shown in Table~\ref{ablation}, while using traditional YUV instead of modified YUV color space in data pre-processing, the MAE increases from 0.75 bpm to 2.35 bpm, and RMSE increases by more than three times. Meanwhile, the r declines by nearly 10\%. The results demonstrate that the modified YUV color space benefits the results of HR estimation using 30s facial videos. The Y channel in the modified YUV color space represents brightness, and U and V represent chromaticity. Brightness refers to the degree of lightness or darkness, which depends on the intensity of the light source and the reflection coefficient of the object's surface. The variation of reflected light intensity in different stages of the heart cycle is reflected in the brightness of the face, so the modified YUV color space pays more attention to the brightness to effectively improve the performance of recovery heart rhythm via facial videos.\par
\subsubsection{Effectiveness of MMSF mechanism}
As shown in Table~\ref{ablation}, while eliminating the MMSF mechanism from the proposed method, the MAE worsens at 2.05 bpm, and the RMSE rises to 4.01 bpm. The r decreases to 0.96. It can be found that MAE and RMSE are more than twice higher than those obtained using the MMSF mechanism. That is, the pyramid-like structure of the multi-scale feature segmentation and fusion in the MMSF mechanism allows the network to learn the high-dimensional features of the signal from multiple scale modes in all directions, taking into account the global and local information of the signals. In addition, signals with complex noise are decomposed into multi-frequency modes, which contain less noise and are more traceable, validating the importance of our proposed MMSF mechanism for improving the robustness of the network.
\subsubsection{Effectiveness of the TMSC and SSA module}
As shown in Table~\ref{ablation}, both MAE and RMSE increase by more than 10\% without the TMSC module. The r decreases a little. The results suggest that the multi-scale convolution kernel in the TMSC module can improve the time-domain field of perception, allowing neurons to base their decisions on more comprehensive information. Furthermore, the multi-scale features extracted by the TMSC module in the time domain have the benefit of improving the accuracy of HR estimation.
We further evaluate the effectiveness of the combination of SSA module. As shown in Table~\ref{ablation}, both MAE and RMSE increase by more than 20\% without using SSA module, while the r decreases a little. This validates our conception that the SSA module can improve the filtering ability of the physiological task against noise. These results show the robustness of the FastBVP-Net, which uses the TMSC module and the SSA module instead of the traditional residual convolution block.
\subsubsection{Effectiveness of oversampling training strategy}
In order to validate the effectiveness of oversampling strategy, we train our network with standard sampling and oversampling strategies. As shown in Table~\ref{ablation}, when putting all training set samples for training in one epoch, the results of HR estimation are worse than using oversampling strategy. The MAE increases from 0.75 to 1.55 bpm, while the RMSE increases from 1.12 to 3.13 bpm. The r reduces to 0.92. It indicates that different heart rate intervals have similar temporal and spatial characteristics. Therefore, equally putting samples of different intervals into the network for training will help the network have better generalization performance and denoising ability in the whole dataset.
\subsection{Analysis of the number of parameters and FLOPs}
Considering the perfect results of CVD~\cite{27CVD2020} and DualGan~\cite{7DualGANJB} in the intra-dataset and cross-dataset experiments, we analyze their detailed structures and compare the number of parameters and floating-point operations per second (FLOPS) with the proposed FastBVP-Net. Since we form the space size of the 2D spatial-temporal map as 4, the proposed FastBVP-Net requires far fewer parameters than existing well-established CVD~\cite{27CVD2020} and DualGan~\cite{7DualGANJB}. The comparative results are illuminated in Table~\ref{paraFLOPs}. In the testing phases, the related module of CVD includes the physiological encoder, non-physiological encoder, the decoder, and the physiological estimator, whose parameters are 42k and FLOPs is $1.1\times10^{10}$. Similarly, the number of parameters in the BVP Estimator of DualGAN is approximately 66k, and FLOPs is $9.9\times10^{9}$. In contrast, the proposed network contains only 11k parameters. The number of parameters is only one-fourth of CVD and one-sixth of DualGAN.  Moreover, the FLOPs of the proposed FastBVP-Net is $1.3\times10^{8}$, which is less than those two comparative networks in order of magnitude. As a result, it is shown that the proposed FastBVP-Net is a lightweight method for measuring heart rhythm with less computation burden.\par 
In the task of monitoring 30 seconds video, the CVD method calculates the average HR of 10-second intervals for the input video. It needs a total of 41 calculations to complete the whole task. A similar situation is observed in DualGAN. On the other hand, the proposed FastBVP-Net can monitor the whole video in one calculation. In this condition, the FastBVP-Net has lower computational complexity than the existing well-established methods. Therefore, the FastBVP-Net could have excellent advantages for the remote measurement of heart rhythm in varied real-world applications.
\begin{center}
\begin{table}[htb]
\centering
\caption{\centering{The number of parameters and FLOPs in current well-established methods}}\label{paraFLOPs}
\setlength{\tabcolsep}{2.5mm}{
\begin{tabular}{ccccc}
\toprule
\textbf{Method} & \textbf{Params} & \textbf{FLOPs}\\ 
\midrule
\textbf{CVD~\cite{27CVD2020}}& $42k$ & $1.1\times10^{10}$\\
\textbf{DualGAN~\cite{7DualGANJB}} & $66k$ & $9.9\times10^{9}$\\
\textbf{FastBVP} & \bm{$11k$} & \bm{$1.3\times10^{8}$} \\
\bottomrule
\end{tabular}}
\end{table} 
\end{center}
\subsection{Performance of shorter videos}
The accurate measurement of heart rhythm via short-time facial videos reflects the network's signal processing capability and robustness. Therefore, we also evaluate the performance of FastBVP-Net for shorter videos (15 seconds), which is tolerable and common in real-world scenarios.\par
In the intra-dataset testing, the performances of the FastBVP-Net in heart rhythm measurements based on 30-second and 15-second videos are compared in Table~\ref{HR-intra} and Table~\ref{HRV-intra}. The FastBVP-Net achieves competitive HR and HRV estimation results using 15-second videos with one calculation. The MAE and RMSE of HR estimation are worse than PulseGAN and DualGAN, which are adopted repeated calculations in 30-second videos. The r is 0.98, which declines slightly compared to the best results obtained via 30-second videos. Moreover, the results of HRV are almost the same as those obtained by FastBVP-Net using 30-second videos. The results are even better than the CVD method using longer videos. Since we calculate the HR based on the reconstructed BVP signal, it is hard to calculate an accurate HR for short-duration signals. The results validate the effectiveness of the proposed method in BVP signal reconstruction using shorter videos.\par
We also evaluate the proposed FastBVP-Net using 15-second videos on MMSE-HR in the cross-dataset testing. Table~\ref{HR-cross} shows that the proposed FastBVP-Net still achieves the competitive results of HR estimation using 15s facial video. Furthermore, compared with other methods using 30s videos, the results of FastBVP-Net (15s) are only worse than CVD and FastBVP-Net (30s). The comparative results on different datasets show that the proposed method performs well for heart rhythm measurements using shorter facial videos.
\section{Conclusion}
It is challenging to measure heart rhythm using rPPG because of the weak BVP signal contaminated by various noises. This paper proposes a novel deep-learning network, FastBVP-Net, to measure heart rhythm quickly with low computational resources. FastBVP-Net uses a multi-frequency mode signal fusion (MMSF) mechanism to suppress noise components and extract stable BVP signals from facial videos. Moreover, an oversampling training strategy is explored to solve the overfitting problem. The proposed MMSF and overfitting sampling scheme can also be applied to other signal-processing tasks. In the future, we will adapt the network to more complex disturbances from varied real-world scenarios.
\section{Acknowledgments}
This work is supported by the Sichuan Science and Technology Program under Grant 2022YFS0032 to Xiujuan Zheng. The authors would like to thank the engineers of Xi'an Singularity Fusion Information Technology Co. Ltd for their supports of this project.

%\section*{CRediT authorship contribution statement}
%\textbf{Jialiang Zhuang}:Conceptualization, Software,Methodolgy, Writing.
%\textbf{Bin Li}:
%\textbf{Xiujuan Zheng}:

%\section*{Declaration of competing interest}
%The authors declare that they have no known competing financial interests or personal relationships that could have appeared to influence the work reported in this paper.\par

\bibliography{mybibfile.bib}
\bibliographystyle{ieeetr}

\end{document}